\documentclass[11pt,a4paper]{article}
\usepackage[hyperref]{naaclhlt2019}
\usepackage{times}
\usepackage{latexsym}
\usepackage{booktabs}
\usepackage{color}
\usepackage{graphicx}
\usepackage[TABBOTCAP]{subfigure}
\usepackage[shortlabels]{enumitem}

\usepackage{url}

\aclfinalcopy 
\def\aclpaperid{526} 

\usepackage{multirow}

\newcommand{\Blue}[1]{\textcolor[rgb]{0.00,0.00,1.00}{#1}}
\newcommand{\Green}[1]{\textcolor[rgb]{0.00,0.7,0.00}{#1}}

\def\CM{{\mathcal C}}

\def\JM{{\mathcal J}}

\def\NM{{\mathcal N}}

\DeclareSymbolFont{extraup}{U}{zavm}{m}{n}
\DeclareMathSymbol{\vardiamond}{\mathalpha}{extraup}{87}

\def\e{{\bf e}}

\def\k{{\bf k}}
\def\o{{\bf o}}

\def\x{{\bf x}}

\def\0{{\bf 0}}
\def\1{{\bf 1}}

\usepackage{amsmath}

\title{
Incorporating Context and External Knowledge \\
for Pronoun Coreference Resolution
}

\author{
Hongming Zhang$^\clubsuit$\thanks{~This work was done during the internship of the first author in Tencent AI Lab.},~ Yan Song$^\spadesuit$,~ and Yangqiu Song$^\clubsuit$ \\
  $^\clubsuit$Department of CSE, The Hong Kong University of Science and Technology \\
  $^\spadesuit$Tencent AI Lab\\
  {\tt hzhangal@cse.ust.hk, clksong@gmail.com, yqsong@cse.ust.hk} \\
}

\date{}

\begin{document}
\maketitle
\begin{abstract}

Linking pronominal expressions to the correct references requires, in many cases, better analysis of the contextual information and external knowledge.
In this paper, we propose a two-layer model for pronoun coreference resolution that leverages both context and external knowledge,
where a knowledge attention mechanism is designed to ensure the model leveraging the appropriate source of external knowledge based on different context.
Experimental results demonstrate the validity and effectiveness of our model, where it outperforms state-of-the-art models by a large margin.



\end{abstract}

\section{Introduction}\label{sec:introduction}





The question of how human beings resolve pronouns has long been of interest to both linguistics and natural language processing (NLP) communities, for the reason that pronoun itself has weak semantic meaning~\cite{ehrlich1981search} and brings challenges in natural language understanding.
To explore solutions for that question, pronoun coreference resolution~\cite{hobbs1978resolving} was proposed.
As an important yet vital sub-task of the general coreference resolution task, pronoun coreference resolution is to find the correct reference for a given pronominal anaphor in the context and has been shown to be crucial for a series of downstream tasks~\cite{mitkov2014anaphora}, including machine translation~\cite{mitkov1995anaphora}, summarization~\cite{steinberger2007two}, information extraction~\cite{edens2003investigation}, and dialog systems~\cite{strube2003machine}.



\begin{figure}[t]
    \centering
    \includegraphics[width=\linewidth]{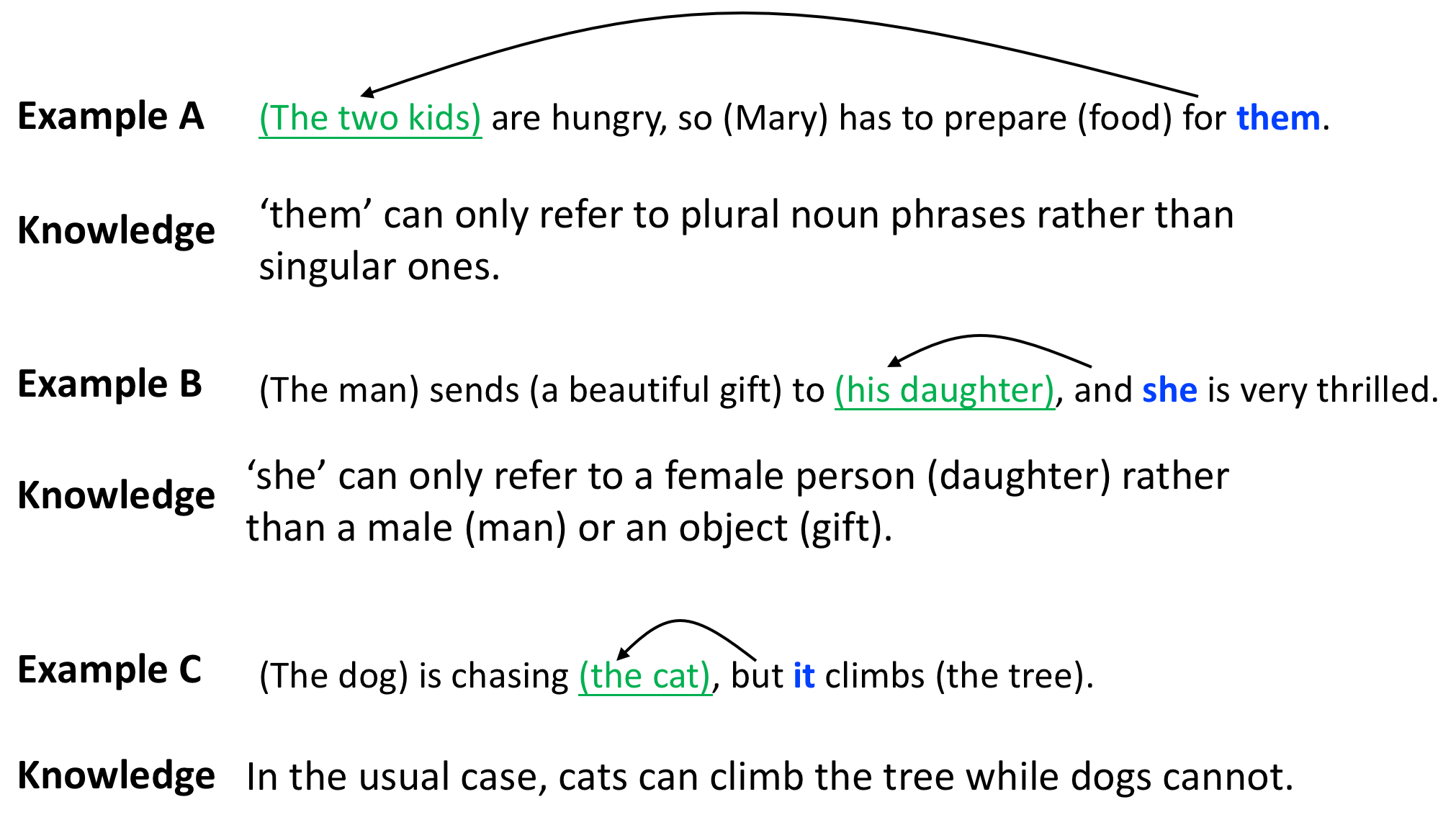}
    \caption{Pronoun coreference examples, where each example requires different knowledge for its resolution.
    Blue bold font refers to the target pronoun,
    where the correct noun reference and other candidates are marked by green underline and brackets, respectively.}
    \label{fig:example}
\end{figure}

Conventionally, people design rules \cite{hobbs1978resolving, nasukawa1994robust, mitkov1998robust} or use features \cite{ng2005supervised, charniak2009works, li2011pronoun} to resolve the pronoun coreferences.
These methods heavily rely on the coverage and quality of the manually defined rules and features.
Until recently, end-to-end solution~\cite{DBLP:conf/emnlp/LeeHLZ17} was proposed towards solving the general coreference problem, where deep learning models were used to better capture contextual information.
However, training such models on annotated corpora can be biased and normally does not consider external knowledge.

Despite the great efforts made in this area in the past few decades~\cite{hobbs1978resolving,mitkov1998robust,ng2005supervised,rahman2009supervised}, pronoun coreference resolution remains challenging. 
The reason behind is that the correct resolution of pronouns can be influenced by many factors~\cite{ehrlich1981search}; many resolution decisions require reasoning upon different 
contextual and external knowledge~\cite{rahman2011coreference}, which is also proved in other NLP tasks \cite{song2017-CoNLL,song2018-NAACL,zhang2018-NAACL}.
Figure \ref{fig:example} demonstrates such requirement with three examples, where
Example A depends on the plurality knowledge that `them' refers to plural noun phrases; 
Example B illustrates the gender requirement of pronouns where `she' can only refer to a female person (girl);
Example C requires a more general type of
knowledge\footnote{This is normally as selectional preference (SP) \cite{hobbs1978resolving}, which is defined as given a predicate (verb), a human has the preference for its argument (subject in this example).} that `cats can climb trees but a dog normally does not'.
All of these knowledge are difficult to be learned from training data.
Considering the importance of both contextual information and external human knowledge, how to jointly leverage them becomes an important question for pronoun coreference resolution.

In this paper, we propose a two-layer model to address the question
while solving two challenges of incorporating external knowledge into deep models for pronoun coreference resolution,
where the challenges include: first, different cases have their knowledge preference, i.e., some knowledge is exclusively important for certain cases, which requires the model to be flexible in selecting appropriate knowledge per case;
second, the availability of knowledge resources is limited and such resources normally contain noise, which requires the model to be robust in learning from them.

Consequently, in our model,
the first layer predicts the relations between candidate noun phrases and the target pronoun based on the contextual information learned by neural networks. 
The second layer compares the candidates pair-wisely, in which we propose a knowledge attention module to focus on appropriate knowledge based on the given context.
Moreover, a softmax pruning is placed in between the two layers to select high confident candidates.
The architecture ensures the model being able to leverage both context and external knowledge.
Especially, compared with conventional approaches that simply treat external knowledge as rules or features, our model is not only more flexible and effective but also interpretable as it reflects which knowledge source has the higher weight in order to make the decision.
Experiments are conducted on a widely used evaluation dataset, where the results prove that the proposed model outperforms all baseline models by a great margin.\footnote{All code and data are available at:\url{https://github.com/HKUST-KnowComp/Pronoun-Coref}.}

Above all, to summarize, this paper makes the following contributions:
\begin{enumerate}[leftmargin=*]
\setlength{\itemsep}{0pt}
\setlength{\parsep}{0pt}
\setlength{\parskip}{0pt}
    \item We propose a two-layer neural model to combine contextual information and external knowledge for the pronoun coreference resolution task.
    \item We propose a knowledge attention mechanism that allows the model to select salient knowledge for different context, which predicts more precisely and can be interpretable through the learned attention scores.
    \item With our proposed model, the performance of pronoun coreference resolution is boosted by a great margin over the state-of-the-art models.
\end{enumerate}

\section{The Task}\label{sec:task_definition}

Following the conventional setting~\cite{hobbs1978resolving}, the task of pronoun coreference resolution is defined as: for a pronoun $p$ and a candidate noun phrase set $\NM$, the goal is to identify the correct non-pronominal references set\footnote{It is possible that a pronoun has multiple references.} $\CM$.
the objective is to maximize the following objective function:
\begin{equation}\label{Objective}
\JM = \frac{\sum_{c \in \CM}{e^{F(c, p)}}}{\sum_{n \in \NM}e^{F(n, p)}},
\end{equation}
where $c$ is the correct reference and $n$ the candidate noun phrase.
$F(\cdot)$ refers to the overall coreference scoring function for each $n$ regarding $p$.
Following~\cite{mitkov1998robust}, all non-pronominal noun phrases in the recent three sentences of the pronoun $p$ are selected to form $N$.

\begin{figure}[t]
    \centering
    \includegraphics[width=\linewidth]{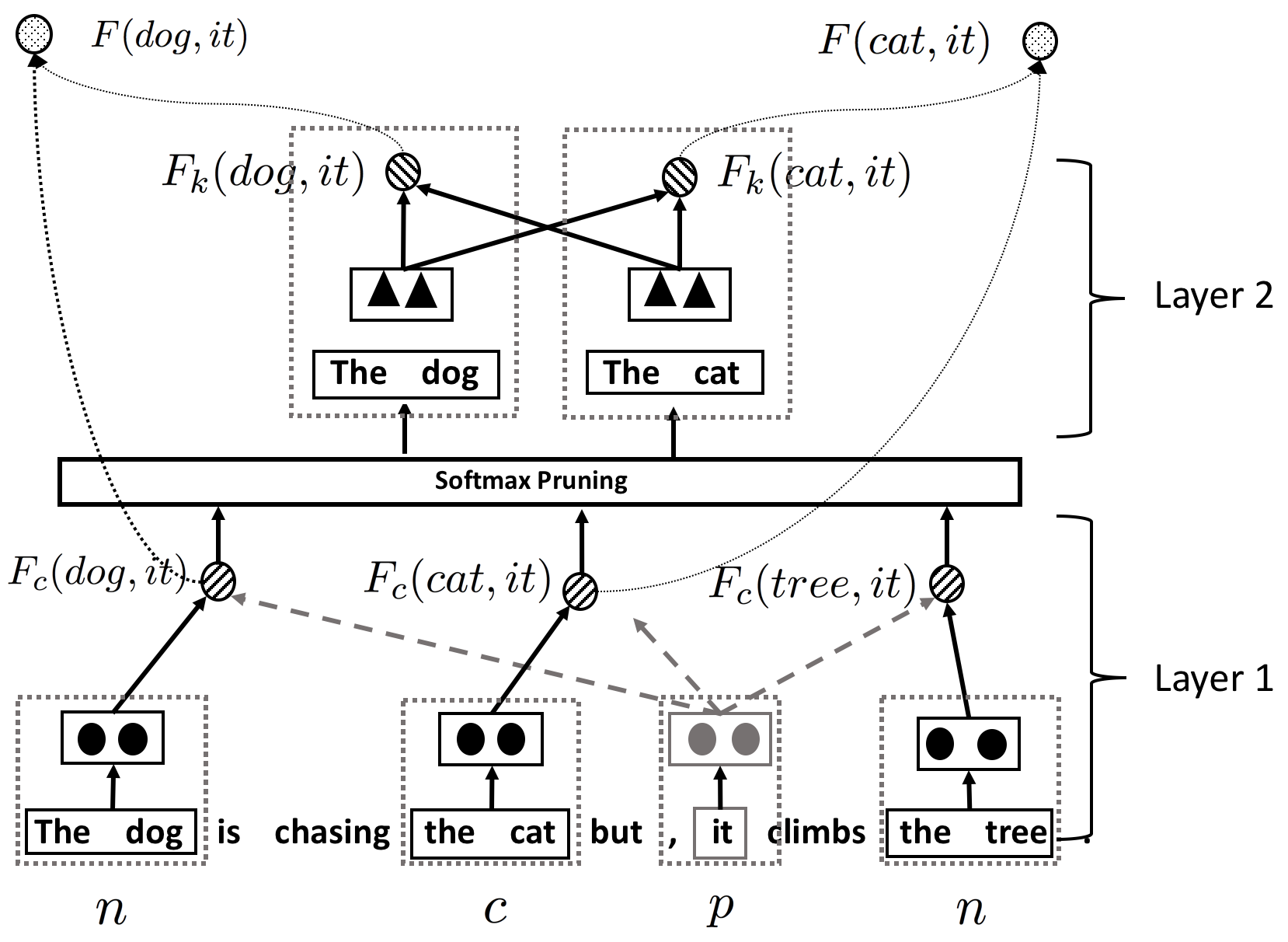}
    \caption{The architecture of the two-layer model for pronoun coreference resolution.
    The first layer encodes the contextual information for computing $F_c$.
    The second layer leverages external knowledge to score $F_k$.
    A pruning layer is applied in between the two layers to control computational complexity.
    The dashed boxes in the first and second layer refer to span representation and knowledge scoring, respectively.
    }
    \label{fig:model}
\end{figure}

Particularly in our setting, we want to leverage
both the local contextual information and external knowledge
in this task,
thus for each $n$ and $p$, $F(.)$ is decomposed into two components:
\begin{equation}\label{score}
	F(n, p) = F_c(n, p) + F_k(n, p),
\end{equation}
where $F_c(n, p)$ is the scoring function that predicts the relation between
$n$ and $p$ based on the contextual information;
$F_k(n, p)$ is the scoring function that predicts the relation between $n$ and $p$ based on the external knowledge.
There could be multiple ways to compute $F_c$ and $F_k$,
where a solution proposed in this paper is described as follows.


\section{The Model}\label{sec:model}



The architecture of our model is shown in Figure~\ref{fig:model}, where we use two layers to incorporate contextual information and external knowledge.
Specifically, the first layer takes the representations of different $n$ and the $p$ as input and predict the relationship between each pair of $n$ and $p$, so as to compute $F_c$.
The second layer leverages the external knowledge to compute $F_k$, which consists of pair-wise knowledge score $f_k$ among all candidate $n$.
To enhance the efficiency of the model, a softmax pruning module is applied to select high confident candidates into the second layer.
The details of the aforementioned components are described in the
following subsections.

\subsection{Encoding Contextual Information}
\begin{figure}[t]
    \centering
    \includegraphics[width=\linewidth]{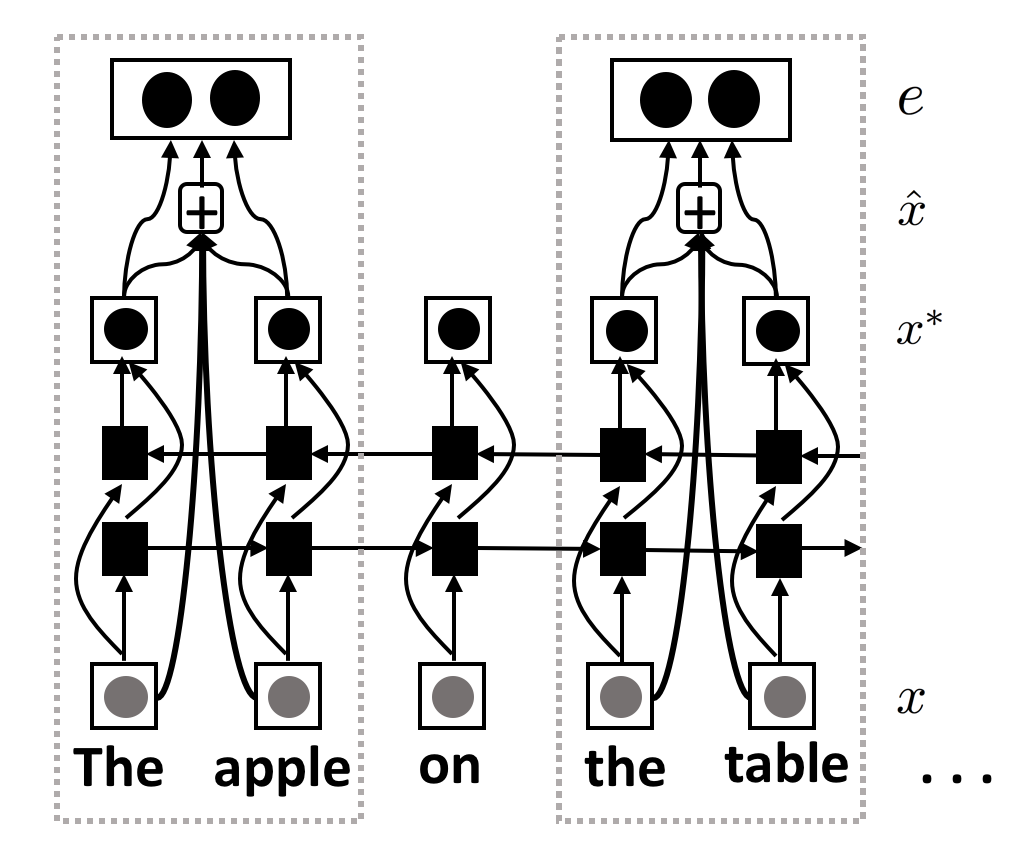}
    \caption{The structure of span representation. Bidirectional LSTM and inner-span attention mechanism are employed to capture the contextual information.}
    \label{fig:mention_representation}
\end{figure}

Before $F_c$ is computed, the contextual information is encoded through a span\footnote{Both noun phrases and the pronoun are treated as spans.} representation (SR) module in the first layer of the model.
Following \citet{DBLP:conf/emnlp/LeeHLZ17}, we adopt the standard bidirectional LSTM (biLSTM) \cite{hochreiter1997long} and the attention mechanism~\cite{bahdanau2014neural} to generate the span representation,
as shown in Figure~\ref{fig:mention_representation}.
Given that the initial word representations in a span $n_i$ are $\x_1,...,\x_T$,

we denote their representations $\x^*_1,...,\x^*_T$ after encoded by the biLSTM.
%
%
Then we obtain the inner-span attention by
\begin{equation}\label{eq:mention_attention}
	a_t = \frac{e^{\alpha_t}}{\sum_{k=1}^{T}e^{\alpha_k}},
\end{equation}
where $\alpha_t$ is computed via a standard feed-forward neural network\footnote{We use $NN$ to present feed-forward neural networks throughout this paper.} $\alpha_t$ = $NN_\alpha(\x^*_t)$.
Thus,
we have the weighted embedding of each span $\hat{x}_i$ through
\begin{equation}\label{eq:overall_embedding}
	\hat{\x}_i = \sum_{k=1}^{T}a_k \cdot \x_k.
\end{equation}
Afterwards, we concatenate the starting ($\x^*_{start}$) and ending ($\x^*_{end}$) embedding of each span, as well as its weighted embedding ($\hat{\x}_i$) and the length feature ($\phi(i)$)
to form its final representation $e$:
\begin{equation}\label{eq:mention_embedding}
	\e_i = [\x^*_{start},\x^*_{end},\hat{\x}_i,\phi(i)].
\end{equation}

Once the span representation of $n \in \NM$ and $p$ are obtained, we compute $F_c$ for each $n$ with a standard feed-forward neural network:
\begin{equation}\label{eq:single_score}
	F_c(n, p) = NN_c([\e_n, \e_p, \e_n \odot \e_p]),
\end{equation}
where
$\odot$ is the element-wise multiplication. 

\subsection{Processing External Knowledge}


In the second layer of our model, external knowledge is leveraged to evaluate all candidate $n$ so as to give them reasonable $F_k$ scores.
In doing so,
each candidate is represented as a group of features from different knowledge sources, e.g., `the cat' can be represented as a singular noun, unknown gender creature, and a regular subject of the predicate verb `climb'.
%
For each candidate, we conduct a series of pair-wise comparisons between it and all other ones to result in its $F_k$ score.
An attention mechanism is proposed to perform the comparison and selectively use the knowledge features.
Consider there exists noise in external knowledge, especially when it is automatically generated, such attention mechanism ensures that, for each candidate, reliable and useful knowledge is utilized rather than ineffective ones.
%
The details of the knowledge attention module and the overall scoring are described as follows.

\begin{figure}[t]
    \centering
    \includegraphics[width=\linewidth]{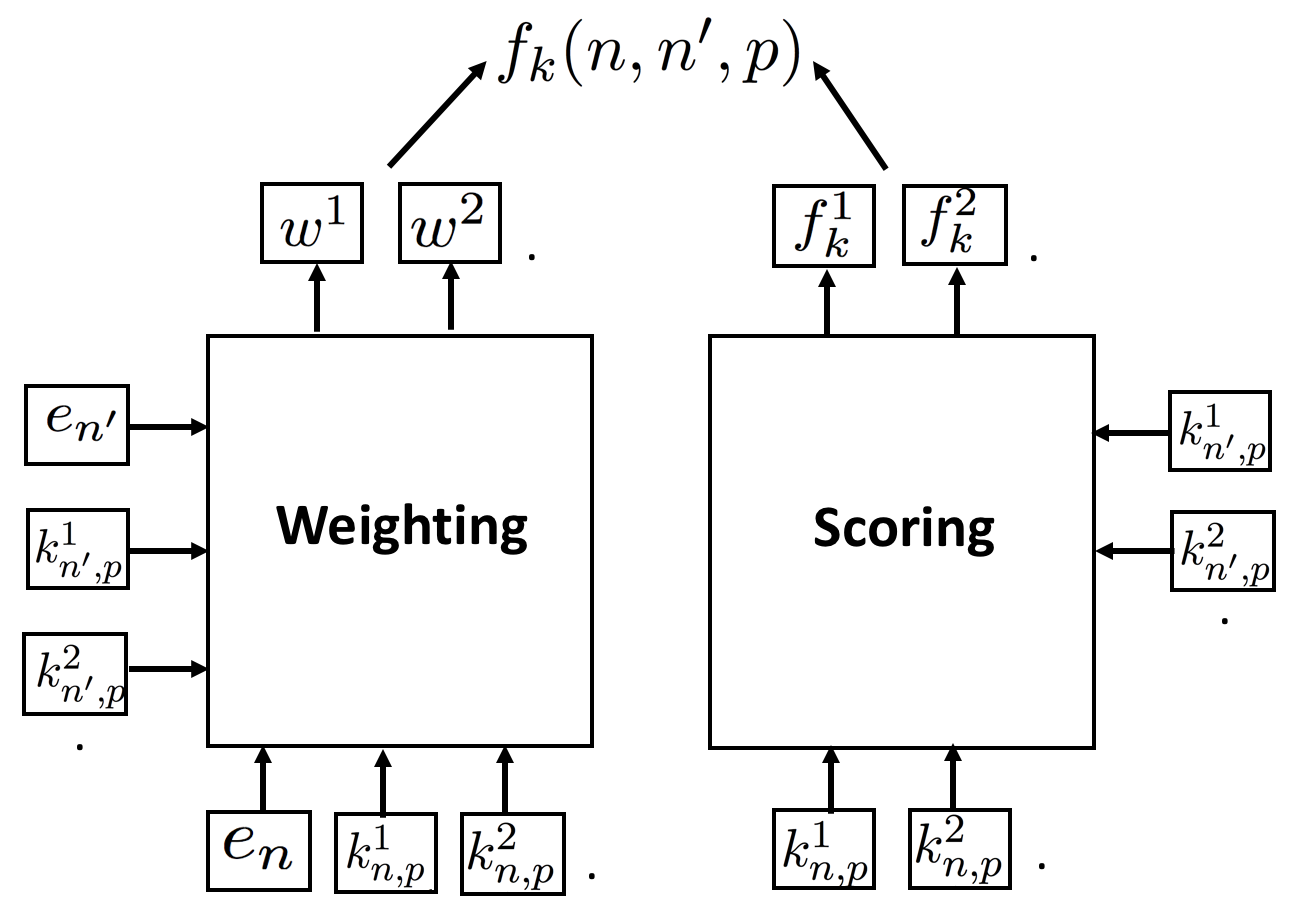}
    \caption{The structure of the knowledge attention module. For each feature $k_i$ from knowledge source $i$, the the weighting component predict its weight $w^i$ and the scoring component computes its knowledge score $f_k^i$. Then a weighted sum is obtained for $f_k$.
    }
    \label{fig:knowledge score}
\end{figure}

\noindent \textbf{Knowledge Attention}
~Figure \ref{fig:knowledge score} demonstrates the structure of the knowledge attention module, where there are two components:
(1) weighting: assigning weights to different knowledge features regarding their importance in the comparison;
(2) scoring: valuing a candidate against another one based on their features from different knowledge sources.
Assuming that there are $m$ knowledge sources input to our model,
each candidate can be represented by $m$ different features, which are encoded as embeddings.
Therefore, two candidates $n$ and $n^\prime$ regarding $p$
have their knowledge feature embeddings
$\k_{n,p}^1, \k_{n,p}^2, ..., \k_{n,p}^m$ and $\k_{n^\prime,p}^1,\k_{n^\prime,p}^2,...,\k_{n^\prime,p}^m$,
respectively.
%
%
%
The weighting component receives all features $\k$ for $n$ and $n^\prime$, and the span representations $\e_n$ and $\e_{n^\prime}$ as input,
where $\e_n$ and $\e_{n^\prime}$ help selecting appropriate knowledge based on the context.
As a result, for a candidate pair ($n$, $n^\prime$) and a knowledge source $i$, its knowledge attention score is computed via
\begin{equation}\label{eq:knowledge_attention_score}
	\beta_i(n, n^\prime, p) = NN_{ka}([\o_{n,p}^i, \o_{n^\prime,p}^i, \o_{n,p}^i \odot \o_{n^\prime,p}^i]),
\end{equation}
where $ \o_{n,p}^i= [\e_n, \k_{n,p}^i]$ and $\o_{n^\prime,p}^i = [\e_{n^\prime}, \k_{n^\prime,p}^i]$ are the concatenation of span representation and external knowledge embedding for candidate $n$ and $n^\prime$ respectively.
%
The weight for features from different knowledge sources is thus computed via
\begin{equation}\label{eq:knowledge_attention_for_one_source}
	w_i = \frac{e^{\beta_i}}{\sum_{j=1}^{m}e^{\beta_j}}.
\end{equation}

Similar to the weighting component, for each feature $i$, we compute its score
$f_k^i(n, n^\prime, p)$ for $n$ against $n^\prime$ in the scoring component through
\begin{equation}\label{eq:knowledge_score_for_one_source}
	f_k^i(n, n^\prime, p) = NN_{ks}([\k_{n,p}^i, \k_{n^\prime,p}^i, \k_{n,p}^i \odot \k_{n^\prime,p}^i]).
\end{equation}
%
where it is worth noting that we exclude $\e$ in this component for the reason that, in practice, the dimension of $\e$ is normally much higher than $\k$. As a result, it could dominate the computation if $\e$ and $\k$ is concatenated.\footnote{We do not have this concern for the weighting component because the softmax (c.f. Eq. \ref{eq:knowledge_attention_for_one_source}) actually amplifies the difference of $\beta$ even if they are not much differentiated.}

Once the weights and scores are obtained,
we have a weighted knowledge score for $n$ against $n^\prime$:
\begin{equation}\label{eq:knowledge_weighted_score}
	f_k(n, n^\prime, p) = \sum_{i=1}^{m}w_i \cdot f_k^i(n, n^\prime, p).
\end{equation}

\noindent\textbf{Overall Knowledge Score}
~After all pairs of $n$ and $n^\prime$ are processed by the attention module, the overall knowledge score for $n$ is computed through the averaged $f_k(n, n^\prime, p)$ over all $n^\prime$:
\begin{equation}\label{eq:knowledge_score}
	F_k(n, p) = \frac{\sum_{n^\prime \in \NM_o} f_k(n, n^\prime, p)}{|\NM_o|},
\end{equation}
where $\NM_o = \NM - n$ for each $n$.


\subsection{Softmax Pruning}
Normally, there could be many noun phrases that serve as the candidates for the target pronoun.
One potential obstacle in the pair-wise comparison of candidate noun phrases in our model is the squared complexity $O(|\NM|^2)$ with respect to the size of $\NM$.
%
To filter out low confident candidates so as to make the model more efficient, we use a softmax-pruning module between the two layers in our model to select candidates for the next step.
The module takes $F_c$ as input for each $n$, uses a softmax computation:
\begin{equation}\label{eq:softmax_pruning}
        \hat{F}_c(n, p) = \frac{e^{F_c(n, p)}}{\sum_{n_i \in \NM}e^{F_c(n_i, p)}}.
\end{equation}
where candidates with higher $\hat{F}_c$ are kept, based on a threshold $t$ predefined as the pruning standard.
Therefore, if candidates have similar $F_c$ scores, the module allow more of them to proceed to the second layer.
Compared with other conventional pruning methods~\cite{DBLP:conf/emnlp/LeeHLZ17,lee2018higher} that generally keep a fixed number of candidates, our pruning strategy is more efficient and flexible.

\section{Experiment Settings}\label{sec:implementation}

\subsection{Data}
 \begin{table}[t]
    \centering
    \small
    \begin{tabular}{l|rrr|r}
        \toprule
        type            & train  & dev   & test  & all    \\
         \midrule
         Third Personal & 21,828 & 2,518 & 3,530 & 27,876 \\
         Possessive     & 7,749  & 1,007 & 1,037 & 9,793  \\
        \midrule
        All     & 29,577 & 3,525 & 4,567 & 37,669 \\
        \bottomrule
    \end{tabular}
    \caption{Statistics of the evaluation dataset. Number of selected pronouns are reported.}
    \label{tab:statics}
\end{table}
The CoNLL-2012 shared task~\cite{pradhan2012conll} corpus is used as the evaluation dataset, which is selected from the Ontonotes 5.0\footnote{https://catalog.ldc.upenn.edu/LDC2013T19}.
Following conventional approaches \cite{ng2005supervised, li2011pronoun}, for each pronoun in the document, we consider candidate $n$ from the previous two sentences and the current sentence.
%
For pronouns, we consider
two types of them following \citet{ng2005supervised}, i.e., third personal pronoun (\textit{she}, \textit{her}, \textit{he}, \textit{him}, \textit{them}, \textit{they}, \textit{it}) and possessive pronoun (\textit{his}, \textit{hers}, \textit{its}, \textit{their}, \textit{theirs}).
Table \ref{tab:statics} reports the number of the two type pronouns and the overall statistics for the experimental dataset.
According to our selection range of candidate $n$, on average, each pronoun has 4.6 candidates and 1.3 correct references.
%


\subsection{Knowledge Types}
In this study, we use two types of knowledge in our experiments.
The first type is linguistic features, i.e., plurality and animacy \& gender. 
We employ the Stanford parser\footnote{https://stanfordnlp.github.io/CoreNLP/}, which generates plurality, animacy, and gender markups for all the noun phrases, to annotate our data.
%
Specifically, the plurality feature denotes each $n$ and $p$ to be singular or plural.
For each candidate $n$, if its plurality status is the same as the target pronoun, we label it 1, otherwise 0.
The animacy \& gender (AG) feature denotes whether a $n$ or $p$ is a living object, and being male, female, or neutral if it is alive.
For each candidate $n$, if its AG feature matches
the target pronoun's, we label it 1, otherwise 0.

The second type is the selectional preference (SP) knowledge.
For this knowledge, we create a knowledge base by counting how many times a predicate-argument tuple appears in a corpus and use the resulted number to represent the preference strength.
Specifically, we use the English Wikipedia\footnote{https://dumps.wikimedia.org/enwiki/} as the base corpus for such counting.
Then we parse the entire corpus through the Stanford parser and record all dependency edges in the format of \textit{(predicate, argument, relation, number)}, where predicate is the governor and argument the dependent in the original parsed dependency edge\footnote{In Stanford parser results, when a verb is a linking verb (e.g., am, is), an 'nsubj' edge is created between its predicative and subject. Thus for this case the predicative is treated as the predicate for the subject (argument) in our study.}. %
%
Later for sentences in the training and test data, we firstly parse each sentence and find out the dependency edge linking $p$ and its corresponding predicate.
Then for each candidate\footnote{If a noun phrase contains multiple words, we use the parsed result to locate its keyword and use it to represent the entire noun phrase.} $n$ in a sentence, we check the previously created SP knowledge base and find out how many times it appears as the argument of different predicates with the same dependency relation (i.e., \textit{nsubj} and \textit{dobj}).
The resulted frequency is grouped into the following buckets [1, 2, 3, 4, 5-7, 8-15, 16-31, 32-63, 64+] and we use the bucket id as the final SP knowledge.
Thus in the previous example:

\vskip 0.5em
\noindent  \textit{The dog} is chasing \textit{the cat} but \textbf{it} climbs \textit{the tree}.
\vskip 0.5em

\noindent Its parsing result indicates that `\textbf{it}' is the subject of the verb `climb'.
Then for `\textit{the dog}', `\textit{the cat}', and `\textit{the tree}', we check their associations with `climb' in the knowledge base and group them in the buckets to form the SP knowledge features.

\begin{table*}[t]
\small
    \centering
    \begin{tabular}{l|ccc|ccc|ccc}
        \toprule
         \multirow{ 2}{*}{Model} & \multicolumn{3}{c|}{Third Personal Pronoun} & \multicolumn{3}{|c|}{Possessive Pronoun} &  \multicolumn{3}{|c}{All}\\
         & P & R & F1 & P & R & F1 & P & R & F1 \\
         
         \midrule

         Recent Candidate & 50.7 & 40.0 & 44.7 & 64.1 & 45.5 & 53.2 & 54.4 & 41.6 & 47.2 \\ 
         Deterministic~\cite{raghunathan2010multi} & 68.7 & 59.4 & 63.7 & 51.8 & 64.8 & 57.6 & 62.3 & 61.0 & 61.7 \\ 
         \midrule

         Statistical~\cite{clark2015entity} & 69.1 & 62.6 & 65.7  & 58.0 & 65.3 & 61.5  & 65.3 & 63.4 & 64.3 \\ 
         Deep-RL~\cite{DBLP:conf/emnlp/ClarkM16} & 72.1 & 68.5 & 70.3  & 62.9 & 74.5 & 68.2 & 68.9 & 70.3 & 69.6 \\ 
         End2end~\cite{lee2018higher} & 75.1 & 83.7 & 79.2 & 73.9 & 82.1 & 77.8 & 74.8 & 83.2 & 78.8 \\ 
         
         \midrule 
         Feature Concatenation & 73.5 & 88.3 & 80.2 & 72.5 & 87.3 & 79.2 & 73.2 & 87.9 & 79.9 \\
         The Complete Model & 75.4 & 87.9 & \textbf{81.2} & 74.9 & 87.2 & \textbf{80.6} & 75.2 & 87.7 & \textbf{81.0} \\
         \bottomrule
    \end{tabular}
    \caption{Pronoun coreference resolution performance of different models on the evaluation dataset. Precision (P), recall (R), and F1 score are reported, with the best one in each F1 column marked bold.
    }
    \label{tab:main_result}
\end{table*}

\subsection{Baselines}
Several baselines are compared in this work. The first two are conventional unsupervised ones:
\begin{itemize}[leftmargin=*]
\setlength{\itemsep}{0pt}
\setlength{\parsep}{0pt}
\setlength{\parskip}{0pt}
\item
\textbf{Recent Candidate}, which simply selects the most recent noun phrase that appears in front of the target pronoun.
\item
\textbf{Deterministic} model \cite{raghunathan2010multi}, which proposes one multi-pass seive model with human designed rules for the coreference resolution task.
\end{itemize}
Besides the unsupervised models, we also compare with three representative supervised ones:
\begin{itemize}[leftmargin=*]
\setlength{\itemsep}{0pt}
\setlength{\parsep}{0pt}
\setlength{\parskip}{0pt}
\item
\textbf{Statistical} model, proposed by \citet{clark2015entity}, uses human-designed entity-level features between clusters and mentions for coreference resolution.
\item
\textbf{Deep-RL} model, proposed by \citet{DBLP:conf/emnlp/ClarkM16}, a reinforcement learning method to directly optimize the coreference matrix instead of the traditional loss function.
\item
\textbf{End2end} is the current state-of-the-art coreference model~\cite{lee2018higher}, which performs in an end-to-end manner and leverages both the contextual information and a pre-trained language model~\cite{peters2018deep}.
\end{itemize}

Note that the Deterministic, Statistical, and Deep-RL models are included in the Stanford CoreNLP toolkit\footnote{https://stanfordnlp.github.io/CoreNLP/coref.html}, and experiments are conducted with their provided code.
For End2end, we use their released code\footnote{https://github.com/kentonl/e2e-coref} and replace its mention detection component with gold mentions for the fair comparison.

To clearly show the effectiveness of the proposed model, we also present a variation of our model as an extra baseline to illustrate the effect of different knowledge incorporation manner:
\begin{itemize}[leftmargin=*]
\setlength{\itemsep}{0pt}
\setlength{\parsep}{0pt}
\setlength{\parskip}{0pt}
\item
\textbf{Feature Concatenation}, a simplified version of the complete model that removes the second knowledge processing layer, but directly treats all external knowledge embeddings as features and concatenates them to span representations.
\end{itemize}


\subsection{Implementation}
Following previous work \cite{lee2018higher}, we use the concatenation of the 300d GloVe embeddings~\cite{pennington2014glove} and the ELMo~\cite{peters2018deep} embeddings as the initial word representations. 
Out-of-vocabulary words are initialized with zero vectors.
Hyper-parameters are set as follows.
The hidden state of the LSTM module is set to 200, and all the feed-forward networks in our model have two 150-dimension hidden layers.
The default pruning threshold $t$ for softmax pruning is set to $10^{-7}$.
All linguistic features (plurality and AG) and external knowledge (SP) are encoded as 20-dimension embeddings. 

For model training, we use cross-entropy as the loss function and Adam \cite{kingma2014adam} as the optimizer.
All the aforementioned hyper-parameters are initialized randomly, and we apply dropout rate 0.2 to all hidden layers in the model.
Our model treats a candidate as the correct reference if its predicted overall score $F(n,p)$
is larger than 0.
The model training is performed with up to 100 epochs, and the best one is selected based on its performance on the development set.

\section{Experimental Results}\label{sec:experiment}


Table~\ref{tab:main_result} compares the performance of our model with all baselines.
Overall, our model performs the best with respect to all evaluation metrics.
Several findings are also observed from the results.
First, manually defined knowledge and features are not enough to cover rich contextual information.
Deep learning models (e.g., End2end and our proposed models), which leverage text representations for context, outperform other approaches by a great margin, especially on the recall.
Second, external knowledge is highly helpful in this task, which is supported by that
our model outperforms the End2end model significantly.

\begin{figure}[t]
    \centering
    \includegraphics[width=\linewidth]{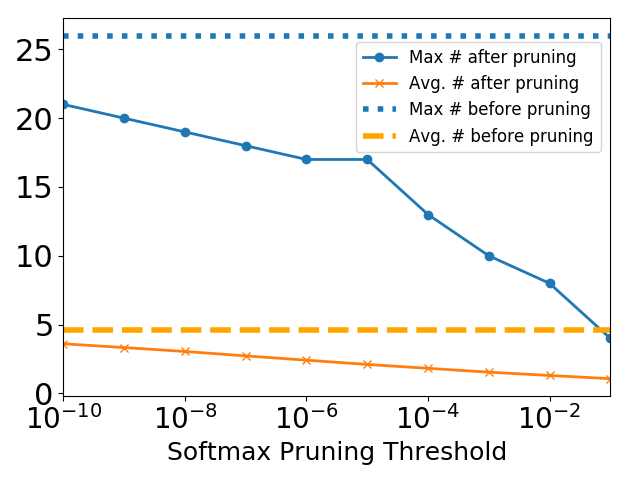}
    \caption{Effect of different thresholds on candidate numbers.
    Max and Average number of candidates after pruning are represented with solid lines in blue and orange, respectively.
    Two dashed lines indicate the max and the average number of candidates before pruning.}
    \label{fig:filtered_candidates}
\end{figure}

Moreover, the comparison between the two variants of our models is also interesting,
where the final two-layer model
outperforms the Feature Concatenation model.
It proves that simply treating external knowledge as the feature, even though they are from the same sources, is not as effective as learning them in a joint framework.
The reason behind this result is mainly from the noise in the knowledge source, e.g., parsing error, incorrectly identified relations, etc.
For example, the plurality of 17\% noun phrases are wrongly labeled in the test data.
As a comparison, our knowledge attention might contribute to alleviate such noise when incorporating all knowledge sources.




\vskip 0.5em
\noindent \textbf{Effect of Different Knowledge} To illustrate the importance of different knowledge sources and the knowledge attention mechanism, we ablate various components of our model and report the corresponding F1 scores on the test data.
The results are shown in Table~\ref{tab:abalation}, which clearly show the necessity of the knowledge.
Interestingly, AG contributes the most among all knowledge types, which indicates that potentially more cases in the evaluation dataset demand on the AG knowledge than others.
More importantly, the results also prove the effectiveness of the knowledge attention module,
which contributes to the performance gap between our model and the Feature Concatenation one.

\begin{table}[t]
    \centering
    \begin{tabular}{l|cc}
        \toprule
         & F1 & $\Delta$F1\\
        \midrule
        The Complete Model  & 81.0 & - \\
        \midrule
        ~--Plurality knowledge & 80.7 & -0.3 \\
        ~--AG knowledge & 80.5 & -0.5 \\
        ~--SP knowledge & 80.4 & -0.6 \\
        \midrule
        ~--Knowledge Attention & 80.1 & -0.9 \\
        \bottomrule
    \end{tabular}
    \caption{Performance of our model with removing different knowledge sources and knowledge attention.
    }
    \label{tab:abalation}
\end{table}

\begin{table*}[t]
    \centering
    \small

    \begin{tabular}{p{2cm}|p{4.5cm}|p{8cm}}
    \toprule
    & Example A & Example B\\
    \midrule
         Sentences & ... (A large group of people) met (Jesus). (A man in the group) shouted to him: ``(Teacher), please come and look at \Green{\underline{(my son)}}. \textbf{\Blue{He}} is the only child I have" ... & ... (My neighbor) told me that there was \Green{\underline{(an accident)}}, and everyone else was intact, except (his father), who was in (hospital) for fractures. I comforted him first and asked (my friend) to rush me to (the hospital). (My neighbor) showed me the police report at (the hospital), which indicated \textbf{\Blue{it}} was all my neighbor's fault. ...\\
         
         \midrule
        Pronoun & \textbf{\Blue{He}} & \textbf{\Blue{it}}\\
        Candidate NPs & A large group of people, Jesus, A man in the group, Teacher, my son.& My friend, an accident, his father, hospital, my friend, the hospital, My neighbor, the hospital.\\
        \midrule
        End2end & Jesus, A man in the group, \Green{\underline{my son}} & None\\
        \midrule
        Our Model & \Green{\underline{my son}} & \Green{\underline{an accident}}\\
        \bottomrule
    \end{tabular}
    \caption{The comparison of End2end and our model on two examples drawn from the test data.
    Pronouns are marked as blue bold font.
    Correct references are indicated in green underline font and other candidates are indicated with brackets.
    `None' refers to that none of the candidates is predicated as the correct reference.}
    \label{tab:case_study}
\end{table*}

\begin{figure}[t]
    \centering
    \includegraphics[width=\linewidth]{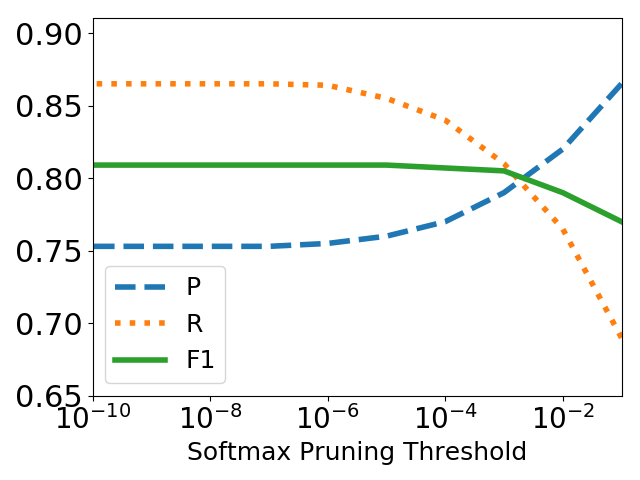}
    \caption{Effect of different pruning thresholds on model performance.
    With the threshold increasing, the precision increases while the recall and F1 drop.}
    \label{fig:pruning}
\end{figure}

\vskip 0.5em
\noindent\textbf{Effect of Different Pruning Thresholds} We try different thresholds $t$ for the softmax pruning in selecting reliable candidates.
The effects of different thresholds on reducing candidates and overall performance are shown in Figure~\ref{fig:filtered_candidates} and \ref{fig:pruning} respectively.
Along with the increase of $t$, both the max and the average number of pruned candidates drop quickly, so that the space complexity of the model can be reduced accordingly.
Particularly, there are as much as 80\% candidates can be filtered out when $t = 10^{-1}$.
Meanwhile, when referring to Figure \ref{fig:pruning}, it is observed that the model performs stable with the decreasing of candidate numbers.
Not surprisingly, the precision rises when reducing candidate numbers, yet the
recall drops dramatically, eventually results in the drop of F1.
With the above observations, the reason we set $t = 10^{-7}$ as the default threshold is straightforward: on this value, one-third candidates are pruned with almost no influence on the model performance in terms of precision, recall, and the F1 score.

\section{Case Study}
\label{sec:case-study}

To further demonstrate the effectiveness of incorporating knowledge into pronoun coreference resolution, two examples are provided for detailed analysis.
The prediction results of the End2end model and our complete model are shown in Table~\ref{tab:case_study}.
There are different challenges in both examples.
In Example A, `Jesus', `man', and `my son' are all similar (male) noun phrases matching the target pronoun `\textit{He}'.
The End2end model predicts all of them to be correct references because their context provides limited help in distinguishing them.
In Example B, the distance between `an accident' and the pronoun `it' is too far. 
As a result, the `None' result from the End2end model indicates that the contextual information is not enough to make the decision.
As a comparison, in our model, integrating external knowledge can help to solve such challenges, e.g., for Example A, SP knowledge helps when Plurality and AG cannot distinguish all candidates.

\begin{figure}
    \centering
    \includegraphics[width=\linewidth]{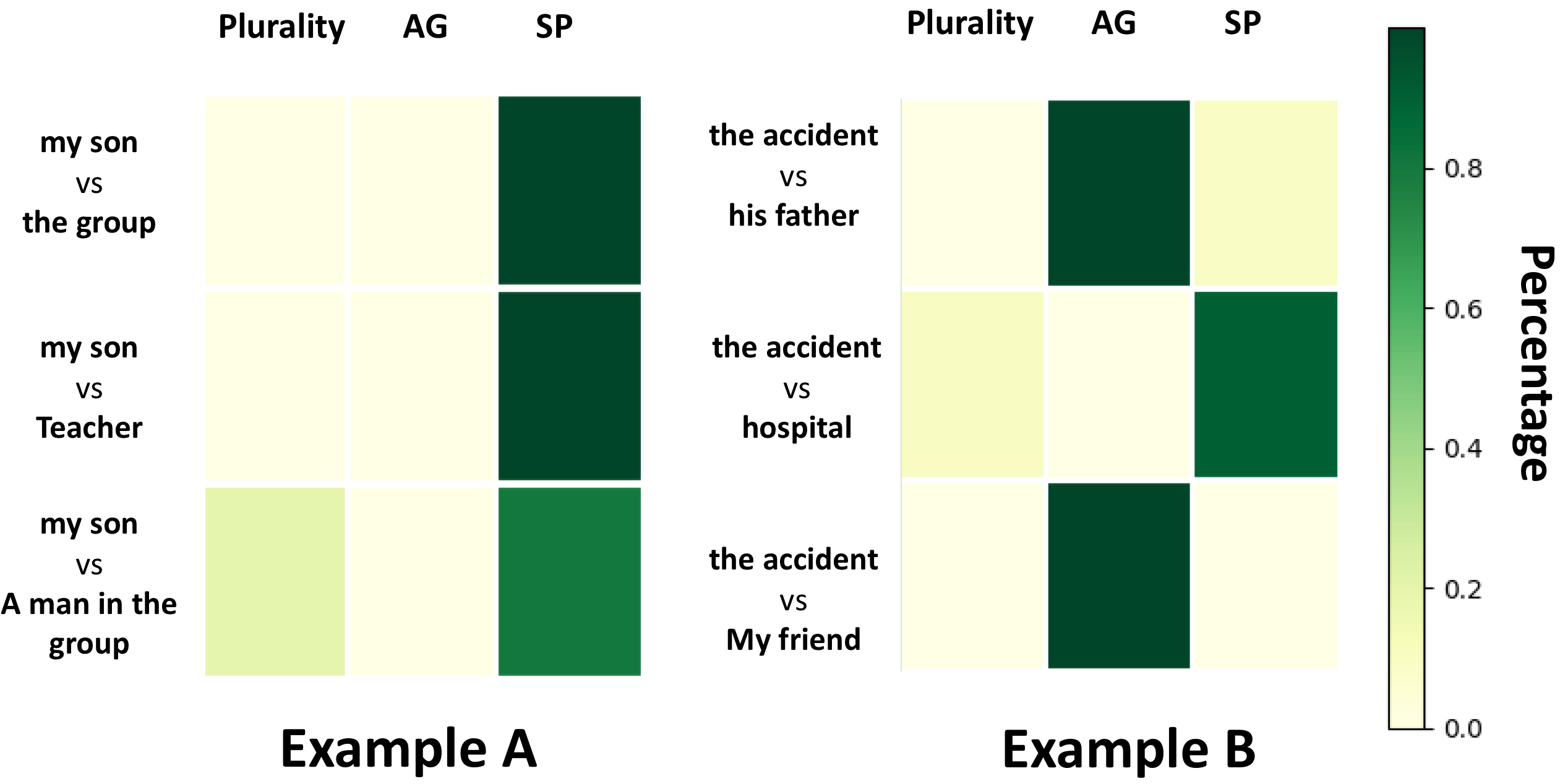}
    \caption{Heatmaps of knowledge attention for two examples, where in each example the knowledge attention weights of the correct references against other candidates are illustrated. Darker color refers to higher weight on the corresponding knowledge type. 
    }
    \label{fig:heat_map}
\end{figure}

To clearly illustrate how our model leverages the external knowledge, we visualize the knowledge attention of the correct reference against other candidates\footnote{Only candidates entered the second layer are considered.} via heatmaps in Figure~\ref{fig:heat_map}.
Two interesting observations are drawn from the visualization.
First, given two candidates, if they are significantly different in one feature, our model tends to pay more attention to that feature.
Take AG as an example,
in Example A, the AG features of all candidates consistently match the pronoun `he' (all male/neutral). Thus the comparison between `my son' and all candidates pay no attention to the AG feature.
While in Example B, the target pronoun `it' cannot describe human, thus 'father' and `friend' are 0 on the AG feature while `hospital' and `accident' are 1.
As a result, the attention module emphasizes AG more than other knowledge types.
%
%
Second, The importance of SP is clearly shown in these examples.
In example A, Plurality and AG features cannot help, the attention module weights higher on SP because `son' appears 100 times as the argument of the parsed predicate `child' in the SP knowledge base, while other candidates appear much less at that position.
In example B, as mentioned above, once AG helps filtering 'hospital' and 'accident',
SP plays an important role in distinguishing them because `accident' appears 26 times in the SP knowledge base as the argument of the `fault' from the results of the parser,
while `hospital' never appears at that position.

\section{Related Work}\label{sec:related-work}
Coreference resolution is a core task for natural language understanding, where it detects mention span and identifies coreference relations among them.
As demonstrated in~\cite{DBLP:conf/emnlp/LeeHLZ17}, mention detection and coreference prediction are the two major focuses of the task.
Different from the general coreference task,
pronoun coreference resolution has its unique challenge since
the semantics of pronouns are often not as clear as normal noun phrases, in general, how to leverage the context and external knowledge to resolve the coreference for pronouns becomes its focus \cite{hobbs1978resolving, rahman2011coreference, emami2018hard}.

In previous work, external knowledge including manually defined rules \cite{hobbs1978resolving, ng2005supervised}, such as number/gender requirement of different pronouns, and world knowledge \cite{rahman2011coreference}, such as selectional preference \cite{wilks1975preferential,DBLP:conf/icmlc2/ZhangS18} and eventuality knowledge~\cite{zhang2019aser}, have been proved to be helpful for pronoun coreference resolution.
Recently, with the development of deep learning, \citet{DBLP:conf/emnlp/LeeHLZ17} proposed an end-to-end model that learns contextual information with an LSTM module and proved that such knowledge is helpful for coreference resolution when the context is properly encoded.
The aforementioned two types of knowledge have their own advantages: the contextual information covers diverse text expressions that are difficult to be predefined while the external knowledge 
is usually more precisely constructed and able to provide extra information beyond the training data.
Different from previous work,
we explore the possibility of joining the two types of knowledge for pronoun coreference resolution rather than use only one of them.
To the best of our knowledge, this is the first attempt that uses deep learning model to incorporate contextual information and external knowledge for pronoun coreference resolution.




\section{Conclusion}
\label{sec:conclusion}
In this paper, we proposed a two-layer model for pronoun coreference resolution,
where the first layer encodes contextual information and the second layer leverages external knowledge.
Particularly, a knowledge attention mechanism is proposed to selectively leverage features from different knowledge sources.
As an enhancement to existing methods, the proposed model combines the advantage of conventional feature-based models and deep learning models, so that context and external knowledge can be synchronously and effectively used for this task.
Experimental results and case studies demonstrate the superiority of the proposed model to state-of-the-art baselines.
Since the proposed model adopted an extensible structure,
one possible future work is to explore the best way to enhance it with more complicated knowledge resources such as knowledge graphs.

\section*{Acknowledgements}
This paper was partially supported by the Early Career Scheme (ECS, No.26206717) from Research Grants Council in Hong Kong. In addition, Hongming Zhang has been supported by the Hong Kong Ph.D. Fellowship and the Tencent Rhino-Bird Elite Training Program.
We also thank 
the anonymous reviewers for their valuable comments and suggestions that help improving the quality of this paper.

\bibliography{resolve_pronoun_with_knowledge}
\bibliographystyle{acl_natbib}

\end{document}